\documentclass[a4paper,twoside]{article}

\usepackage{epsfig}
\usepackage{subcaption}
\usepackage{calc}
\usepackage{amssymb}
\usepackage{amstext}
\usepackage{amsmath}
\usepackage{amsthm}
\usepackage{multicol}
\usepackage{pslatex}
\usepackage{algorithm2e}
\usepackage{booktabs}
\usepackage{pgfplots}
\usepackage{pgfplotstable}
\usepackage{fontawesome}
\usepackage{colortbl}
\usepackage{multirow}
\pgfplotsset{compat=1.18}
\usepackage[bottom]{footmisc}
\usepackage{natbib}
\usepackage{hyperref}
\hypersetup{
    colorlinks=True,
    linkcolor=black,
    citecolor=black,
    filecolor=black,
    urlcolor=magenta,
    pdfborder={0 0 0}
}
\usepackage{tikz}
\usetikzlibrary{external} 
\usepackage{SCITEPRESS}     
\DeclareMathAlphabet\mathbfcal{OMS}{cmsy}{b}{n}

\newcommand{\cam}[1]{{\fontsize{8}{16}\selectfont \color{brown} #1}}

\definecolor{blue}{cmyk}{0.50,0.15,0.0,0.0}
\newcommand{\graycell}{\cellcolor{gray!10}}
\newcommand{\bluecell}{\cellcolor{blue!20}}

\begin{document}

\title{HandMvNet: Real-Time 3D Hand Pose Estimation Using Multi-View Cross-Attention Fusion}

\author{\authorname{Muhammad Asad Ali\sup{1,2}, Nadia Robertini\sup{1} and Didier Stricker\sup{1,2}}
\affiliation{\sup{1}Augmented Vision Group, German Research Center for Artificial Intelligence (DFKI), Kaiserslautern, Germany}
\affiliation{\sup{2}Department of Computer Science, University of Kaiserslautern-Landau (RPTU), Kaiserslautern, Germany}
\email{\{firstname\_middlename.lastname\}@dfki.de}
}


\abstract{In this work, we present HandMvNet, one of the first real-time method designed to estimate 3D hand motion and shape from multi-view camera images. Unlike previous monocular approaches, which suffer from scale-depth ambiguities, our method ensures consistent and accurate absolute hand poses and shapes. This is achieved through a multi-view attention-fusion mechanism that effectively integrates features from multiple viewpoints. In contrast to previous multi-view methods, our approach eliminates the need for camera parameters as input to learn 3D geometry. HandMvNet also achieves a substantial reduction in inference time while delivering competitive results compared to the state-of-the-art methods, making it suitable for real-time applications. Evaluated on publicly available datasets, HandMvNet qualitatively and quantitatively outperforms previous methods under identical settings. Code is available at \textit{\href{https://github.com/pyxploiter/HandMvNet}{github.com/pyxploiter/HandMvNet}}.}

\onecolumn \maketitle \normalsize \setcounter{footnote}{0} \vfill

\section{\uppercase{Introduction}}
\label{sec:introduction}

3D hand pose estimation has emerged as a important research area in computer vision with applications across fields like augmented reality (AR), virtual reality (VR), and robotics. The ability to accurately capture and reconstruct hand movements holds immense potential in enhancing human-computer interaction, enabling more natural, intuitive gesture-based controls. In AR and VR, realistic and responsive hand pose estimation enriches immersive experiences, allowing users to interact seamlessly with virtual environments. Similarly, in robotics, precise hand pose estimation is important for tasks such as robotic hand retargeting, where robotic hands mimic human movements to perform intricate tasks. 

\begin{figure}[th]
 \centering
 \includegraphics[width=0.45\textwidth]{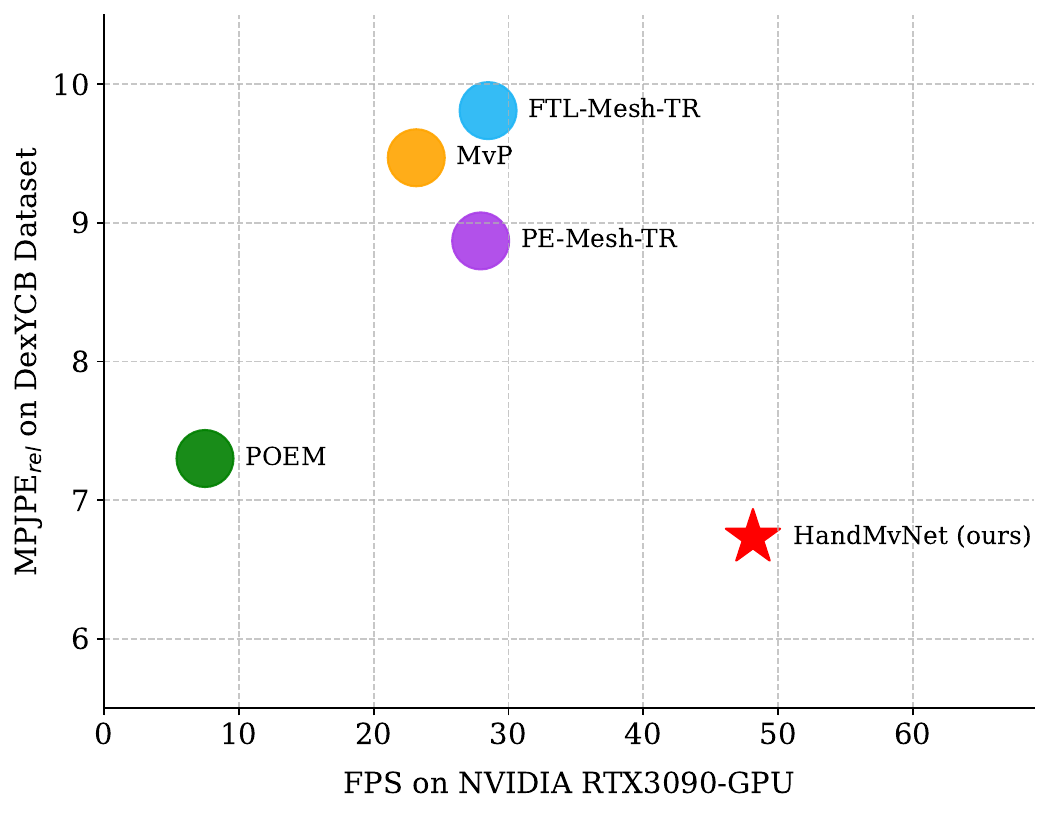}
 \caption{Comparison of error vs. inference speed across different methods. Our approach outperforms other methods in both inference speed and accuracy.}
 \label{fig:fps}
\end{figure}

Traditional approaches in 3D hand pose estimation have primarily relied on single-view images \citep{boukhayma20193d, chen2021i2uv, chen2021camera, ge20193d, moon2020i2l, park2022handoccnet}. However, 3D hand pose estimation from monocular views presents several challenges. Depth and scale ambiguity, where the exact distance and size of the hand from the camera are difficult to determine, significantly complicates the estimation process. Consequently, many approaches only estimate root-relative hand vertices \citep{moon2020i2l, ge20193d, zhou2020monocular}. Occlusions, caused by the overlapping of fingers or the hand being partially obscured by other objects, further add to the complexity of accurately estimating hand poses \citep{park2022handoccnet}. Additionally, varying perspectives and unknown camera viewpoints introduce uncertainties that make the task more challenging. 

To address the limitations associated with monocular views, multi-view setups have been proposed as a solution \citep{yu2021local, chao2021dexycb, yang2022oakink, hampali2020honnotate}. A multi-view setup, consisting of multiple cameras positioned at different angles around the hand, can significantly reduce the impact of occlusions and depth ambiguities, enabling more accurate and robust estimation of hand poses and shapes at absolute 3D locations. 
Most multi-view approaches \citep{Guan2006MultiviewA3, yang2023poem, zheng2023hamuco} are computationally expensive, primarily due to the increased input space and architectural design choices that prioritize qualitative results over computational efficiency. 

In this work, we propose HandMvNet, a novel neural network architecture for efficient and accurate 3D hand pose estimation from multi-view inputs. The key contributions of this work are as follows:

\begin{itemize}
    \item We present a framework that leverages multi-view data for accurate 3D hand pose estimation.
    
    \item Our method achieves real-time performance, making it suitable for time-critical applications.
    
    \item We show that our approach performs effectively with or without camera calibration.
\end{itemize}

We conduct extensive experiments on public multi-view datasets for hand pose and shape reconstruction in challenging scenarios, including strong occlusions from object interactions. Our findings demonstrate that HandMvNet effectively and accurately estimates hand poses and shapes, outperforming existing state-of-the-art methods both qualitatively and computationally.

\begin{figure*}[t]
  \vspace{-0.2cm}
  \centering
   {\includegraphics[width=1\linewidth]{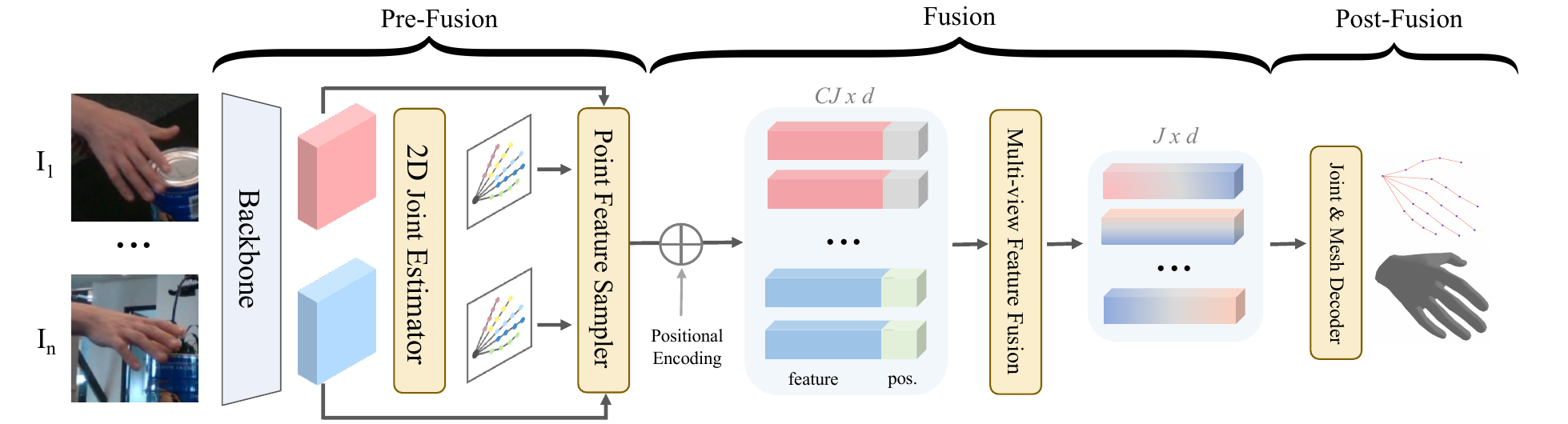}}
  \caption{HandMvNet' architecture consists of three stages: (a) Sampling joint-aligned features using predicted 2D joints from each image (b) Fusing multi-view sampled features, (c) Regressing 3D hand joints and vertices.}
  \label{fig:pipeline}
\end{figure*}

\begin{figure*}[t]
    \centering
    \begin{subfigure}[c]{0.3\linewidth}
        \centering
        \includegraphics[width=\linewidth]{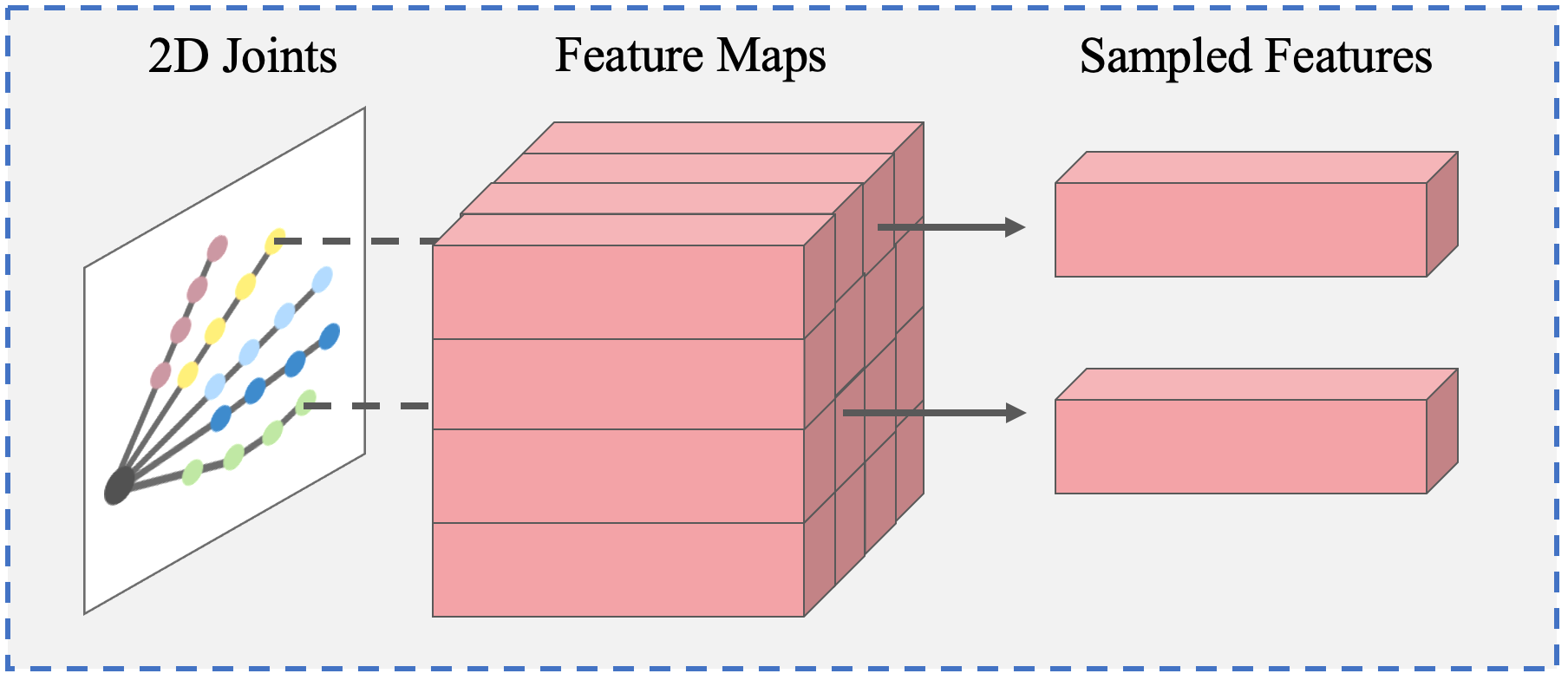}
        \caption{}
        \label{fig:pfs}
    \end{subfigure}
    \hspace{0.005\linewidth}
    \begin{subfigure}[c]{0.29\linewidth}
        \centering
        \includegraphics[width=\linewidth]{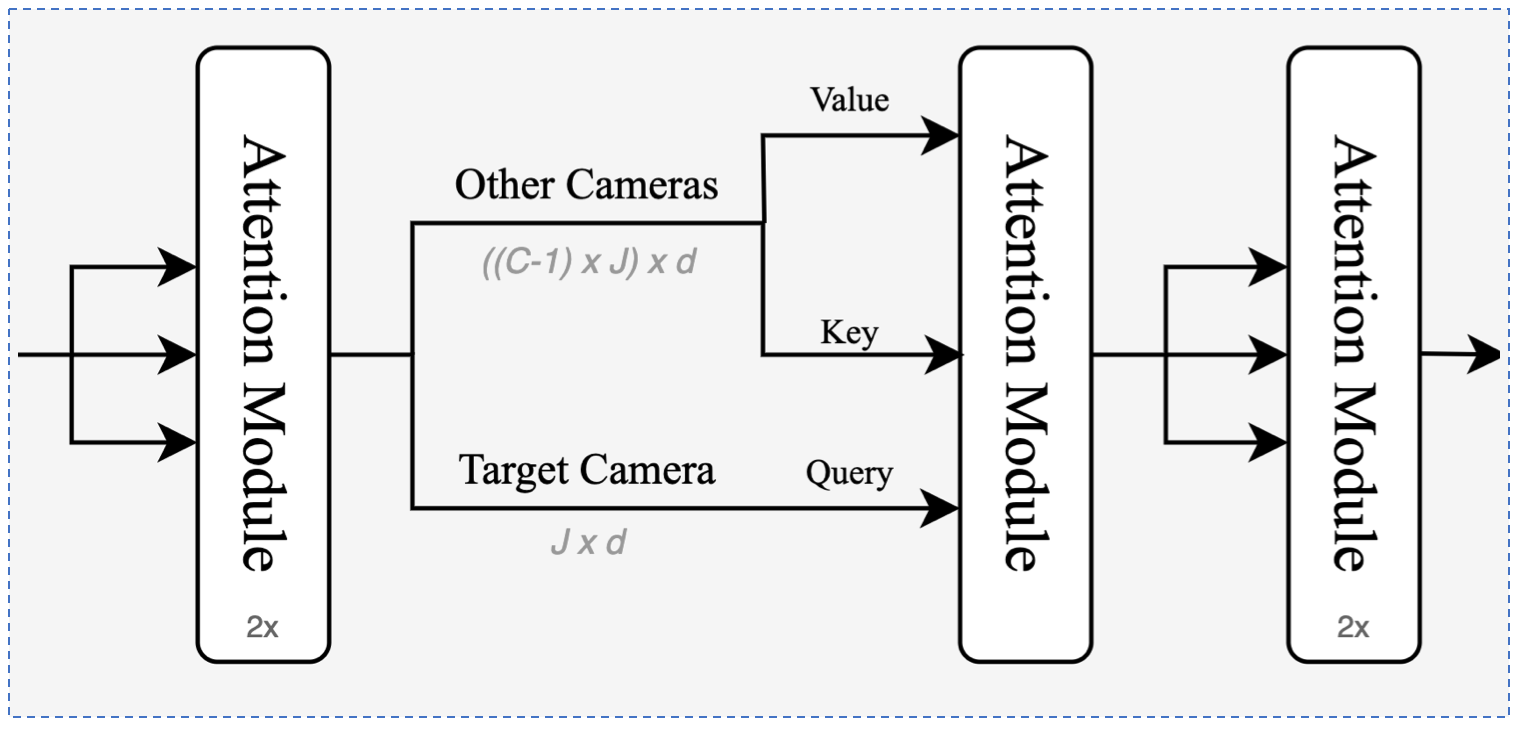}
        \caption{}
        \label{fig:mvff}
    \end{subfigure}
    \hspace{0.005\linewidth}
    \begin{subfigure}[c]{0.135\linewidth}
        \centering
        \includegraphics[width=\linewidth]{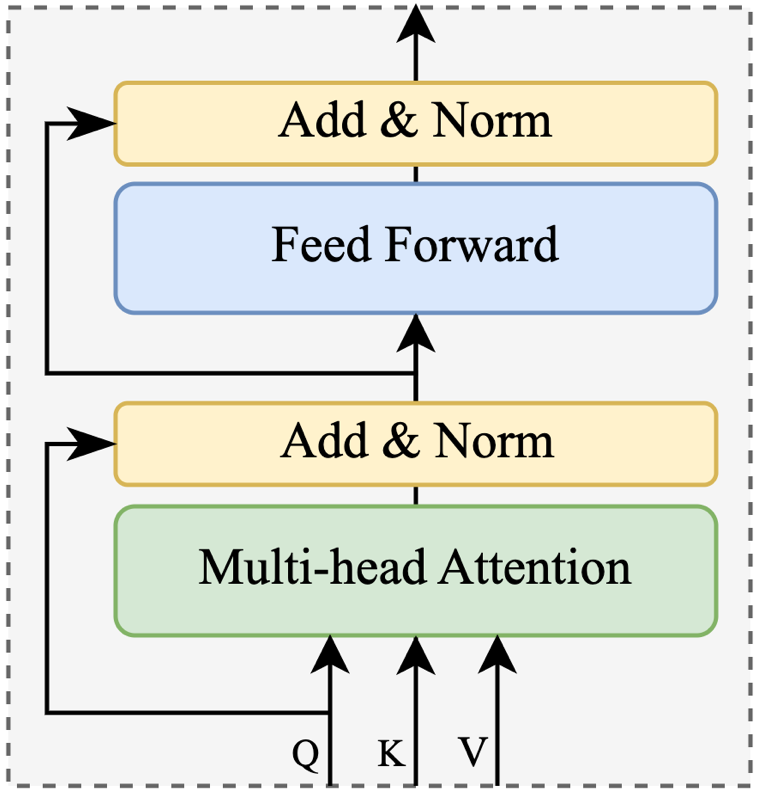}
        \caption{}
        \label{fig:attn}
    \end{subfigure}
    \hspace{0.005\linewidth}
    \begin{subfigure}[c]{0.2\linewidth}
        \centering
        \includegraphics[width=\linewidth]{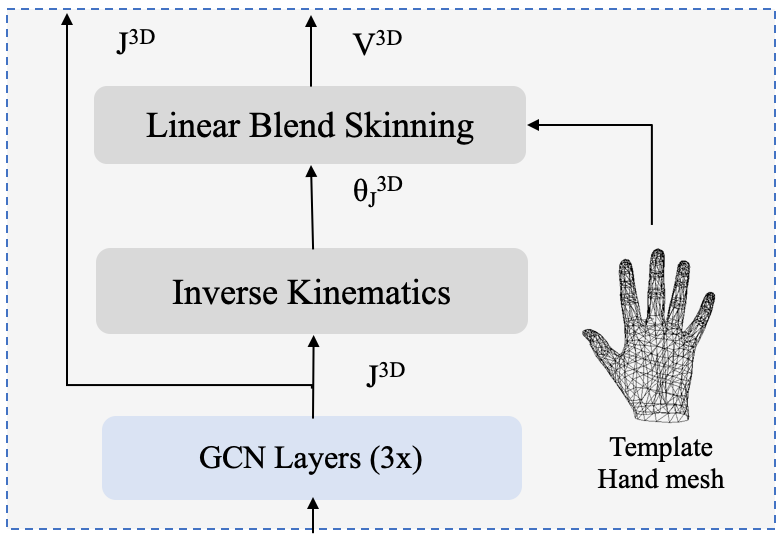}
        \caption{}
        \label{fig:jmd}
    \end{subfigure}
    \caption{Modules of HandMvNet's architecture: (a) Point Feature Sampler. (b) Multi-view Feature Fusion. (c) Attention Module. (d) Joint \& Mesh Decoder}
    \label{fig:modules}
\end{figure*}

\section{\uppercase{Related Work}}

    Most approaches have focused on estimating hand pose from monocular input
    \citep{ge20193d, boukhayma20193d, zhou2020monocular, chen2021i2uv, chen2021camera, park2022handoccnet, moon2020i2l}.
    While various hand representations have been proposed \citep{chen2021i2uv,malik2020handvoxnet, malik2021handvoxnet++}, the deformable hand mesh model MANO \citep{romero2022embodied}, which includes dense 3D hand surface representation, remains the most widely used \citep{chen2021i2uv, park2022handoccnet, zhou2020monocular, boukhayma20193d}.
    Similarly to \citep{ge20193d}, we uniquely estimate the hand mesh directly, bypassing the need for the MANO model parameters, thus offering a flexible, model-free solution. With the rise of transformer architectures \citep{vaswani2017attention}, such frameworks have also been adopted for 3D pose estimation, showcasing their effectiveness \citep{park2022handoccnet, zhao2022graformer, lin2021end}. Despite recent advances, most methods focus on estimating root-relative hand poses due to limited input information and scale-depth ambiguity. In this work, we integrate contributions from multiple views using cross-attention, enabling the estimation of contextualized 3D absolute hand poses.
    Compared to other multi-view approaches \citep{ge2016robust, he2020epipolar, han2022umetrack, remelli2020lightweight, iskakov2019learnable}, our method avoids conventional volumetric or other intermediate representations that negatively affect the inference speed.
    Although most approaches require multi-view camera calibration, mainly for algebraic triangulation and geometric priors to estimate 3D hand pose \citep{remelli2020lightweight, bartol2022generalizable, chen2022structural,iskakov2019learnable, tu2020voxelpose, he2020epipolar, zhang2021adafuse}, we instead propose a more flexible, calibration-free solution that can optionally incorporate camera parameters. Recent advancements \citep{yang2023poem, shuai2022adaptive, ma2022ppt} in transformer-based implicit cross-view fusion inspire our proposed method for multi-view cross-attention fusion.
    
\section{\uppercase{Method}}
\label{sec:method}
The aim of our HandMvNet approach is to estimate 3D hand joints and vertices from multi-view RGB images. In this section, we provide a comprehensive description of our proposed model architecture.

    \subsection{Architecture}
    The overall pipeline of HandMvNet is illustrated in Figure \ref{fig:pipeline}. The network processes a set of multi-view RGB images, $\mathbfcal{I} = \{\mathbf{I}_i\}_{i=1}^{\mathcal{C}}$, captured from $\mathcal{C}$ camera views and estimates the 3D hand joints $\mathbf{J}^{3D} \in \mathbb{R}^{\mathcal{J} \times 3}$ and vertices $\mathbf{V}^{3D} \in \mathbb{R}^{\mathcal{V} \times 3}$, where $\mathcal{J} = 21$ \& $\mathcal{V} = 778$. 

    The architecture consists of three key stages: \textbf{(1) Pre-Fusion:} Each input image is independently processed to extract view-specific features and estimate 2D joint locations, with shared network weights across all views. \textbf{(2) Fusion:} The extracted features are then fused to aggregate multi-view information for enhanced 3D understanding (see Figure \ref{fig:mvff}). \textbf{(3) Post-Fusion:} Finally, the fused features are refined to regress the 3D hand joints and vertices, producing the complete 3D hand reconstruction. Each stage is described in detail in the sections below.

    \subsubsection{Pre-Fusion}
    \noindent \textbf{Backbone:}
    The first stage of our pipelines uses ResNet50 \citep{he2016deep} as a backbone to extract the view-specific image features from input images. The backbone is pre-trained on the ImageNet dataset \citep{imagenet_cvpr09}, and its weights are shared across each camera view. For each camera view \(i\), the backbone processes the image \(\mathbf{I}_i\) and outputs a corresponding view-specific feature map \( \mathbf{Z}_i \in \mathbb{R}^{1024 \times 32 \times 32} \). 

    \medskip \noindent \textbf{2D Joint Estimator:}
    At this stage, two convolutional layers refine the features \( \mathbf{Z}_i \) to produce joint-specific heatmaps $\mathbf{H}_i$. To extract the 2D joint locations from the heatmaps, we apply a differentiable soft-argmax function \citep{sun2018integral}, which transforms the heatmaps into directly usable joint coordinates \( \mathbf{J}^{2D}_i = \textit{soft-argmax}(f_{\text{CNN}}(\mathbf{Z}_i)) \in \mathbb{R}^{\mathcal{J} \times 2} \).

    \medskip \noindent \textbf{Point Feature Sampler:}
    In the final pre-fusion stage, we extract view-specific features from $\mathbf{Z}_i$ (see Figure \ref{fig:pfs}), reduced to a dimensionality of $\mathbb{R}^{512 \times 32 \times 32}$ using a convolutional layer, corresponding to 2D joint locations $\mathbf{J}^{2D}_i$, $\mathbf{S}_i = \textit{sampler}(\mathbf{Z}_i, \mathbf{J}^{2D}_i), \quad \mathbf{S}_i \in \mathbb{R}^{\mathcal{J} \times 512}$.
    The sampled joint-aligned features from all camera views are
    concatenated, forming the aggregated multi-view feature representation \( \mathbf{S} = \textit{concat}(\mathbf{S}_1, \mathbf{S}_2, \dots, \mathbf{S}_{\mathcal{C}})\), where \(\mathbf{S} \in \mathbb{R}^{\mathcal{C} \mathcal{J} \times 512} \).
 
    \subsubsection{Fusion}
    \noindent\textbf{Positional Encoding:} To preserve critical spatial and geometric information in cropped hand images, we introduce three types of positional encodings:
    
    \noindent 1) $\text{PE}_{\text{joint}} \in \mathbb{R}^{\mathcal{CJ} \times 2}$ embeds 2D joint positions into the feature vector to capture the hand's skeletal structure and the relative joint positions in each view. 
    
    \noindent 2) $\text{PE}_{\text{crop}} \in \mathbb{R}^{\mathcal{C} \times 10}$ encodes the location of the hand crop relative to the camera \citep{prakash2023mitigating}, with each corner and one center point $(x,y)$ calculated as $\theta_x = \tan^{-1}((x - p_x)/f_x)$ and $\theta_y = \tan^{-1}((y - p_y)/f_y)$, where $p_x$, $p_y$ are the principal point coordinates, and $f_x$, $f_y$ are focal lengths. $\text{PE}_{\text{crop}}$ is repeated $\mathcal{J}$ times for each joint in the view. This encoding is only applied if camera intrinsics are available.
    
    \noindent 3) Sinusoidal encoding $\text{PE}_{\text{sin}} \in \mathbb{R}^{\mathcal{CJ} \times d}$ \citep{vaswani2017attention} captures inter-view and inter-joint relations for attention-based fusion.

    The final feature vector is: 
    \begin{equation}
        \mathbf{F} = \text{concat}(\mathbf{S}, \text{PE}_{\text{joint}}, \text{PE}_{\text{crop}}) + \text{PE}_{\text{sin}}.
    \end{equation}
    where $\mathbf{F} \in \mathbb{R}^{\mathcal{CJ} \times d}$ and $d = 512+2+10 = 524$. 
        
    \medskip \noindent \textbf{Multi-view Feature Fusion:}
    To capture the dependencies between non-local joints across $\mathcal{C}$ camera views, we pass the independently sampled features $\mathbf{F}$ through attention module (Figure \ref{fig:attn}) and then, to fuse multi-view features, we employ multi-head cross-attention between the first camera view features \( \mathbf{F}_1 \in \mathbb{R}^{\mathcal{J} \times d} \) acting as the \textbf{query} and the features from the remaining camera views \( \mathbf{F}_{\mathcal{C}-1} \in \mathbb{R}^{((\mathcal{C}-1) \times \mathcal{J}) \times d} \) acting as the \textbf{key} and \textbf{value} (source).
    The cross-attention is formulated as:

    \begin{equation}
        \mathbf{F}^* = \textit{softmax} \left( \frac{\mathbf{F}_1 \mathbf{F}_{\mathcal{C}-1}^T}{\sqrt{d}} \right) \mathbf{F}_{\mathcal{C}-1}
    \end{equation}
    The cross-attention layer outputs \( \mathbf{F}^* \in \mathbb{R}^{\mathcal{J} \times d} \) where $\mathcal{J} = 21$ and $d = 524$, which aggregates the features across the camera views into the target camera feature space.
    Finally, self-attention is applied again to $\mathbf{F}^*$ to further refine the intra-joint relationships.

\begin{table*}[ht]
    \caption{Quantitative results (mm) on the test sets of DexYCB-MV, HO3D-MV, and MVHand. \cam{\faCamera} denotes the methods that require camera parameters. The best and second-best results are highlighted in bold and underlined respectively. }

    \label{tab:quantitative}
    \centering
    \footnotesize
    \renewcommand{\arraystretch}{1.05}
    \resizebox{\linewidth}{!}{ 
    \begin{tabular}{c|c|l|ccc|ccc}
    \hline
    & \# & Methods & MPJPE$_{rel}\downarrow$ & PA$_{J}\downarrow$ & AUC$_{J@20}\uparrow$ & MPVPE$_{rel}\downarrow$ & PA$_{V}\downarrow$ & AUC$_{V@20}\uparrow$ \\ \hline
    \multirow{9}{*}{\rotatebox[origin=c]{90}{DexYCB-MV}} 
    & \texttt{1} & \cam{\faCamera} MvP               & 9.47  & 4.26  & \textbf{0.69} & 12.18 & 8.14  & 0.53 \\
    & \texttt{2} & \cam{\faCamera} PE-Mesh-TR         & 8.87  & 4.76  & 0.64 & 8.67  & 4.70  & 0.64 \\
    & \texttt{3} & \cam{\faCamera} FTL-Mesh-TR        & 9.81  & 5.51  & 0.59 & 9.80  & 5.75  & 0.59 \\
    & \texttt{4} & \cam{\faCamera} POEM       & 7.30  & \textbf{3.93}  & \underline{0.68} & \underline{7.21}  & \textbf{4.00}  & \textbf{0.70} \\
    & \texttt{5} & \cam{\faCamera} Multi-view Fit. & 8.77  & 5.19  & 0.65 & 8.71  & 5.29  & \underline{0.65} \\
    & \graycell \texttt{6} & \graycell \cam{\faCamera} HandMvNet (ours) & \graycell \textbf{6.73}  & \graycell \underline{4.08}  & \graycell 0.67 & \graycell \textbf{7.19}  & \graycell \underline{4.52}  & \graycell \underline{0.65} \\ 
    & \graycell \texttt{7} & \graycell \cam{\faCamera} HandMvNet-HR (ours) & \graycell \underline{6.89}  & \graycell \underline{4.08}  & \graycell 0.67 & \graycell 7.30 & \graycell 4.53 & \graycell \underline{0.65} \\ 
    & \bluecell \texttt{8} & \bluecell HandMvNet w/o cam. (ours)  & \bluecell 7.03  & \bluecell 4.13  & \bluecell 0.66 & \bluecell 7.38  & \bluecell 4.56  & \bluecell 0.64 \\
    & \bluecell \texttt{9} & \bluecell HandMvNet-HR w/o cam. (ours)  & \bluecell 7.28  & \bluecell 4.20 & \bluecell 0.65 & \bluecell 7.62  & \bluecell 4.69 & \bluecell 0.63 \\ \hline
      
    \multirow{9}{*}{\rotatebox[origin=c]{90}{HO3D-MV}}
     & & & & & AUC$_{J@50}\uparrow$ & & & AUC$_{V@50}\uparrow$ \\
    & \texttt{10} & \cam{\faCamera} MvP               & 24.90  & 10.44  & 0.60 & 27.08 & 10.04  & 0.59 \\
    & \texttt{11} & \cam{\faCamera} PE-Mesh-TR         & 30.23  & 11.67  & 0.54 & 29.19  & 11.31  & 0.55 \\
    & \texttt{12} & \cam{\faCamera} FTL-Mesh-TR        & 34.74  & 10.72  & 0.52 & 33.53  & 10.56  & 0.53 \\
    & \texttt{13} & \cam{\faCamera} POEM       & 21.94  & \textbf{9.60}  & \textbf{0.63} & 21.45  & \underline{9.97}  & \textbf{0.66} \\
    & \graycell \texttt{14} & \graycell \cam{\faCamera} HandMvNet (ours)     & \graycell 21.43  & \graycell 10.89 & \graycell 0.59 & \graycell 20.17 & \graycell 10.16 & \graycell 0.61 \\
    & \graycell \texttt{15} & \graycell \cam{\faCamera} HandMvNet-HR (ours)   & \graycell \underline{20.73}  & \graycell 11.01  & \graycell \underline{0.61} & \graycell \underline{19.82} & \graycell 10.73 & \graycell 0.62 \\
    & \bluecell \texttt{16} & \bluecell HandMvNet w/o cam. (ours) & \bluecell 21.55  & \bluecell \underline{10.15}  & \bluecell 0.58 & \bluecell 20.10 & \bluecell \textbf{9.39} & \bluecell 0.61 \\
    & \bluecell \texttt{17} & \bluecell HandMvNet-HR w/o cam. (ours) & \bluecell \textbf{20.40}  & \bluecell 11.98  & \bluecell \underline{0.61} & \bluecell \textbf{19.33} & \bluecell 11.24 & \bluecell \underline{0.63} \\ \hline
    
    \multirow{6}{*}{\rotatebox[origin=c]{90}{MVHand}}
     & & & & & AUC$_{J@20}\uparrow$ & & & AUC$_{V@20}\uparrow$ \\
    & \texttt{18} & \cam{\faCamera} MediaPipe-DLT       & 17.24 & 9.97 & 0.28 & 18.42 & 7.74 & 0.21 \\ 
    & \graycell \texttt{19} & \graycell \cam{\faCamera} HandMvNet (ours)     & \graycell 2.07 & \graycell 1.30 & \graycell \underline{0.90} & \graycell \underline{7.57} & \graycell 4.14 & \graycell \underline{0.62} \\
    & \graycell \texttt{20} & \graycell \cam{\faCamera} HandMvNet-HR (ours)   & \graycell \underline{1.86} & \graycell \underline{1.21} & \graycell \textbf{0.91} & \graycell 7.59 & \graycell \underline{4.12} & \graycell \underline{0.62} \\
    & \bluecell \texttt{21} & \bluecell HandMvNet w/o cam. (ours) & \bluecell 2.05  & \bluecell 1.28 & \bluecell \underline{0.90} & \bluecell 7.62 & \bluecell \textbf{4.11} & \bluecell \underline{0.62} \\
    & \bluecell \texttt{22} & \bluecell HandMvNet-HR w/o cam. (ours)   & \bluecell \textbf{1.77} & \bluecell \textbf{1.14} & \bluecell \textbf{0.91} & \bluecell \textbf{7.46} & \bluecell 4.15  & \bluecell \textbf{0.63} \\ \hline
    \end{tabular}
    } 
\end{table*}

    \subsubsection{Post-Fusion} \label{sec:post-fusion}
    \noindent \textbf{Joint \& Mesh Decoder:} 
    We use a three-layer graph convolutional network (GCN) to decode 3D joints from the fused feature $\mathbf{F}^* \in \mathbb{R}^{\mathcal{J} \times d}$, treating $\mathcal{J}$ joints as graph nodes with $d$-dimensional features, estimating the final $\mathbf{J}^{3D} \in \mathbb{R}^{\mathcal{J} \times 3}$. Inverse Kinematics (IK) is then applied to compute joint rotation angles $\theta_{J^{3D}} \in \mathbb{R}^{(\mathcal{J}-5) \times 3}$, which form a hand skeleton. This skeleton deforms a hand template mesh via linear blend skinning to yield the final 3D vertices $\mathbf{V}^{3D} \in \mathbb{R}^{\mathcal{V} \times 3}$ as shown in Figure \ref{fig:jmd}. 

    \subsection{Training}
    
    We apply mean squared error loss for the predicted 2D heatmaps ($L_{\text{H}}$) and L1 loss for both 2D and 3D joints ($L_{\text{2D}}$, $L_{\text{3D}}$). Additionally, if camera parameters are available, we project predicted 3D joints onto 2D camera views using the perspective function $\Pi_{c}(\cdot): \mathbb{R}^3 \rightarrow \mathbb{R}^2$, and minimize the L1 loss between these projections and the ground-truth 2D joints ($L_{\text{G2D}}$), as well as the predicted 2D joints ($L_{\text{P2D}}$). The total loss is defined as:
    \begin{align}
        L &= \lambda_{\text{H}}L_{\text{H}}
            + \lambda_{\text{2D}}L_{\text{2D}}
            + \lambda_{\text{3D}}L_{\text{3D}}\notag \\ &\quad
            \textcolor{gray}{+ \lambda_{\text{G2D}}L_{\text{G2D}}} 
            \textcolor{gray}{+ \lambda_{\text{P2D}}L_{\text{P2D}}}
    \end{align}
    where $\lambda$ values are set as 10, 1, 1, 1, and 0.5 to balance the loss scale, respectively.

\section{\uppercase{Experiments and Results}}
In this section, we conduct experiments to validate and assess the effectiveness of our proposed architecture, along with providing implementation details.
We use Pytorch \citep{paszke2019pytorch} to implement all our networks. The AdamW \citep{loshchilov2017decoupled} optimizer is used with a weight decay of 0.05 and an initial learning rate set to 0.0001. The model is trained on two RTXA6000 GPUs with a batch size of 32. Cropped hand images resized to 256×256, serve as input data. We also evaluate a variation of our model, denoted as HandMvNet-HR, which uses HRNet-w40 as backbone \citep{sun2019deep}.

\begin{table*}[thbp]
    \centering
    \caption{Ablation Studies.}
    \label{tab:ablation}
    \vspace{0.1cm}
    \begin{subtable}{0.35\textwidth}
        \caption{Different positional encodings.}
        \label{tab:pos-enc}
        \centering
        \resizebox{\linewidth}{!}{
        \begin{tabular}{l|cccc}
            \hline
            Pos. Encoding & MPJPE$_{rel}\downarrow$  & PA$_{J}\downarrow$ & AUC$_{J}\downarrow$  \\
            \hline
            sin & 7.69 & 4.40 & 0.63 \\
            sin + joint & 6.96 & 4.14 & 0.66 \\
            sin + joint + crop & \textbf{6.73} & \textbf{4.08} & \textbf{0.67} \\
            \hline
        \end{tabular}}
    \end{subtable}
    \hfill
    \begin{subtable}{0.32\textwidth} 
        \caption{Effect of fusion layers}
        \label{tab:fusion-layers}
        \centering
        \resizebox{\linewidth}{!}{
        \begin{tabular}{c|ccc}
            \hline
            Fusion Layers & MPJPE$_{rel}$$\downarrow$  & PA$_{J}$$\downarrow$  & AUC$_{J}$$\uparrow$  \\
            \hline
            3      & 6.90 & 4.16 & 0.66 \\
            5      & \textbf{6.73} & \textbf{4.08} & \textbf{0.67} \\
            7      & 6.88 & 4.14 & 0.67 \\
            \hline
        \end{tabular}}
    \end{subtable}%
    \hfill
    \begin{subtable}{0.32\textwidth} 
        \caption{Different number of camera views}
        \label{tab:num-cams}
        \centering 
        \resizebox{\linewidth}{!}{
        \begin{tabular}{c|ccc}
            \hline
            Camera views & MPJPE$_{rel}$$\downarrow$  & PA$_{J}$$\downarrow$  & AUC$_{J}$$\uparrow$  \\
            \hline
            8           & \textbf{6.73} & \textbf{4.08} & \textbf{0.67} \\
            4           & 7.47 & 4.38 & 0.64 \\
            2           & 8.33 & 4.83 & 0.60 \\
            \hline
        \end{tabular}}
    \end{subtable}
\end{table*}

\begin{figure*}[htb]
    \centering
    \begin{subfigure}[b]{0.32\textwidth}
        \centering
        \includegraphics[width=\textwidth]{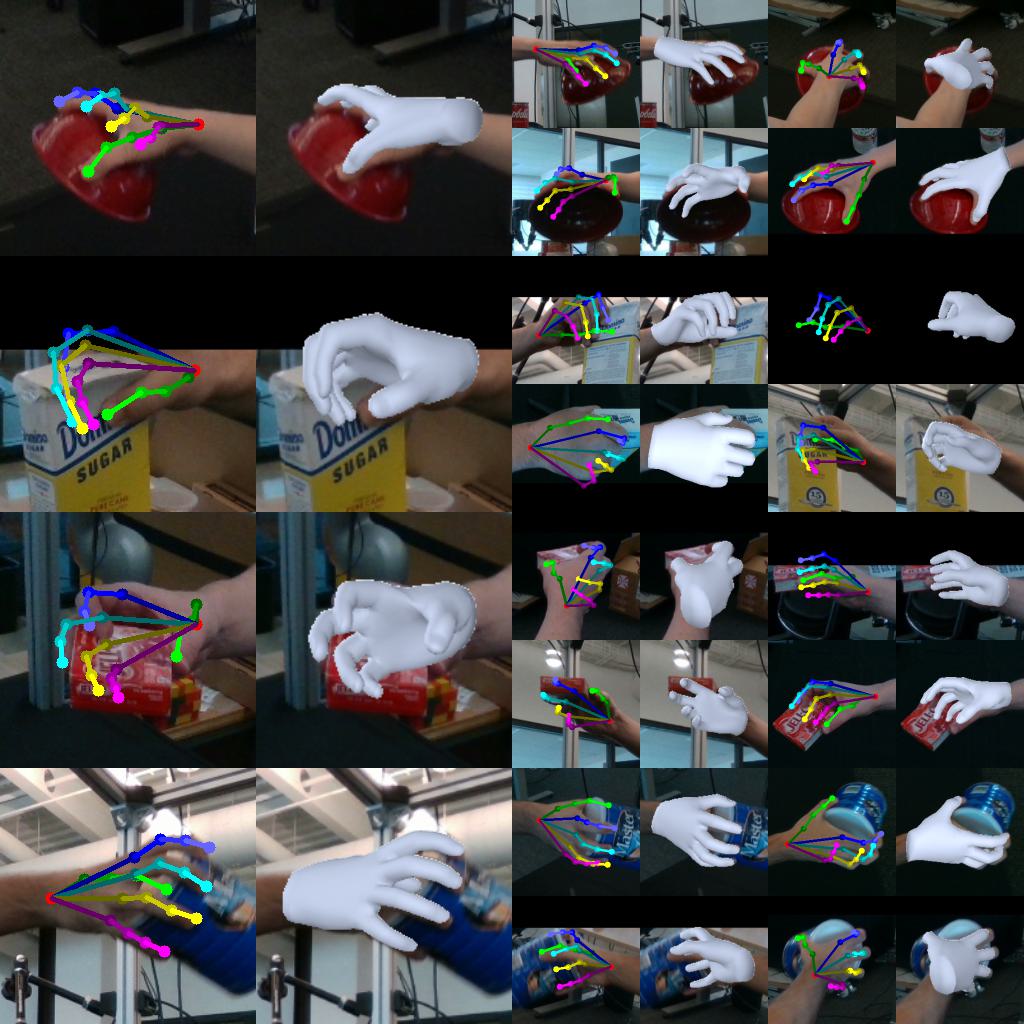}
    \end{subfigure}
    \hfill
    \begin{subfigure}[b]{0.32\textwidth}
        \centering
        \includegraphics[width=\textwidth]{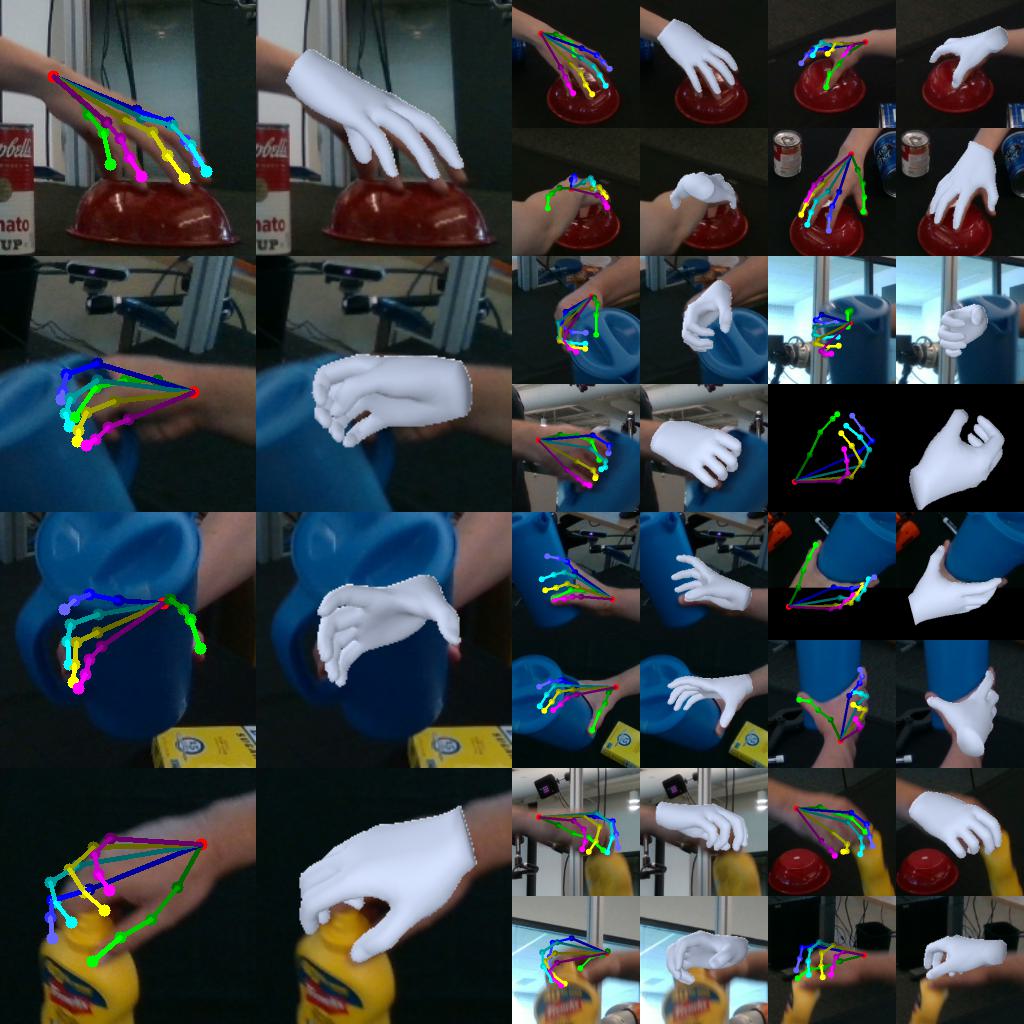}
    \end{subfigure}
    \hfill
    \begin{subfigure}[b]{0.32\textwidth}
        \centering
        \includegraphics[width=\textwidth]{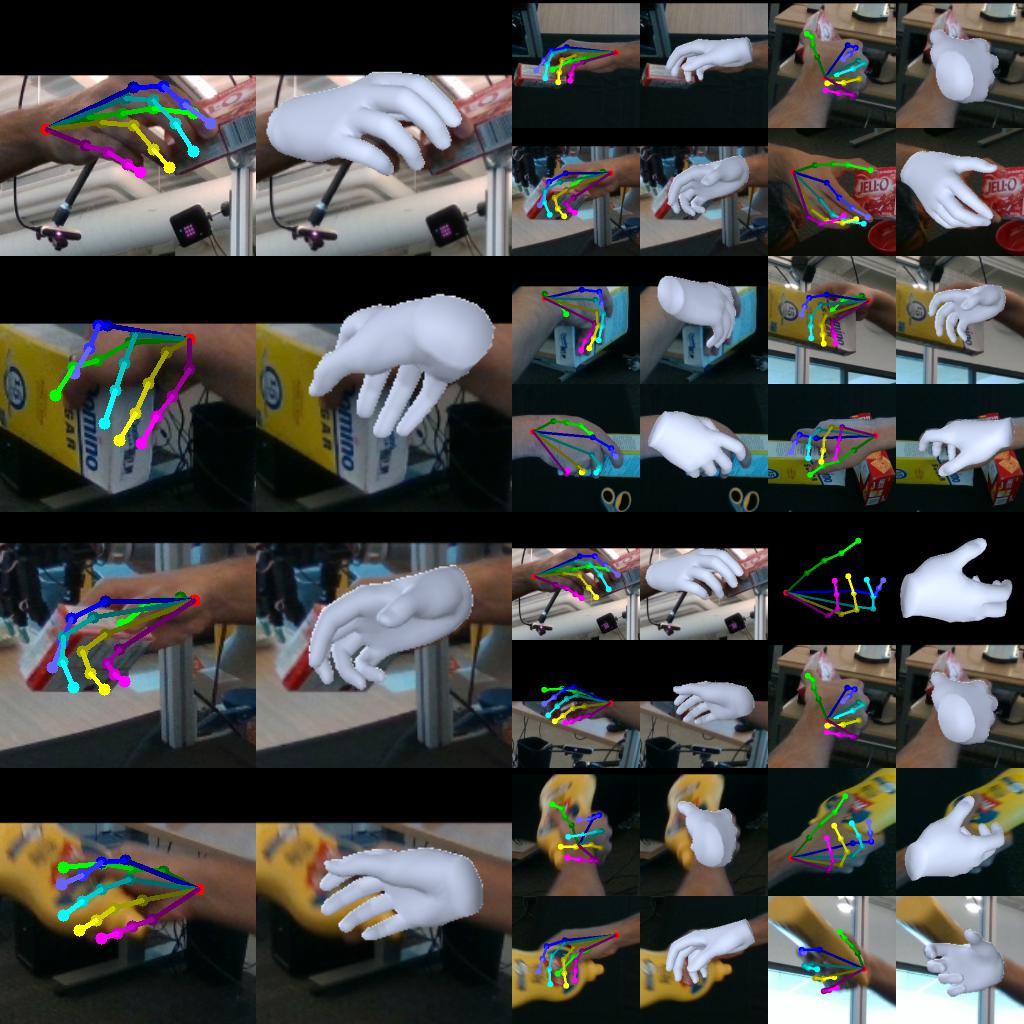}
    \end{subfigure}
    \caption{Qualitative results on the test set of DexYCB-MV dataset.}
    \label{fig:dexycb_vis}
\end{figure*}

\subsection{Datasets}
    \noindent \textbf{DexYCB} \citep{chao2021dexycb} is a multi-view RGB-D dataset capturing hand-object interactions, featuring 10 subjects and 8 camera views per subject. We follow the official ``S0" split, excluding left-hand samples, resulting in 25,387 training, 1,412 validation, and 4,951 test multi-view samples, same as \citep{yang2023poem}. We refer to this split as DexYCB-MV.
    
    \smallskip \noindent \textbf{HO3D} (v3) \citep{hampali2020honnotate} includes images of hand-object interaction from up to 5 cameras. We construct HO3D-MV by selecting 7 sequences with complete multi-view observations from all 5 cameras. For the training set, we use the sequences `ABF1',`BB1', `GSF1', `MDF1', and `SiBF1', while the sequences `GPMF1' and `SB1' are reserved for testing. This results in 9,087 training and 2,706 test multi-view samples.
    
    \smallskip \noindent \textbf{MVHand} \citep{yu2021local} is a multi-view RGB-D hand pose dataset featuring 4 subjects and 4 camera views per subject. We split the 21,200 multi-view frames into 15,417 training, 1,927 validation, and 3,856 test multi-view samples.

\subsection{Evaluation Metrics}

We evaluate the performance of our method using the following standard hand pose estimation metrics. 
\textbf{1) MPJPE$_{rel}$/MPVPE$_{rel}$} (Mean Per Joint/Vertex Position Error) calculates the average Euclidean distance (in mm) between predicted and ground-truth joints/vertices, after aligning the root(-wrist) joint. 
\textbf{2) PA-MPJPE/PA-MPVPE} (Procrustes Aligned Joint/Vertex Error) measures MPJPE/MPVPE after applying procrustes analysis for scale, center and rotation alignment. We refer to these metrics as PA$_J$ and PA$_V$ in our experiments.
\textbf{3) AUC$_J$/AUC$_V$} (Area Under Curve for Joint/Vertex Error) computes the area under the percentage of correct keypoints (PCK) curve over a range of thresholds.

\subsection{Comparison with Previous Methods}
We benchmark our 3D hand reconstruction approach against state-of-the-art (SOTA) multi-view methods, including \textbf{POEM} \citep{yang2023poem} and \textbf{MvP} \citep{zhang2021direct}. Although MvP is primarily designed for multi-person pose estimation, we focus on its performance in single-hand reconstruction. Given the limited availability of multi-view hand pose methods, we further evaluate simulated approaches that combine single-view hand reconstruction with advanced multi-view fusion techniques. Detailed descriptions of these simulated methods, such as \textbf{PE-Mesh-TR} \citep{liu2022petr,lin2021end}, \textbf{FTL-Mesh-TR} \citep{remelli2020lightweight}, and \textbf{Multi-view Fitting} \citep{hampali2020honnotate}, are provided in Section 4.2 of \citep{yang2023poem}. For the MVHand dataset, which lacks established multi-view benchmarks, we introduce a baseline "Mediapipe-DLT" that estimates 2D joints using Mediapipe \citep{zhang2020mediapipe}, triangulates them via Direct Linear Transform (DLT) \citep{hartley2003multiple}, and obtains 3D vertices through linear blend skinning.

\begin{figure*}[htb]
    
    \centering
    \begin{subfigure}[b]{0.32\textwidth}
        \centering
        \includegraphics[width=\textwidth]{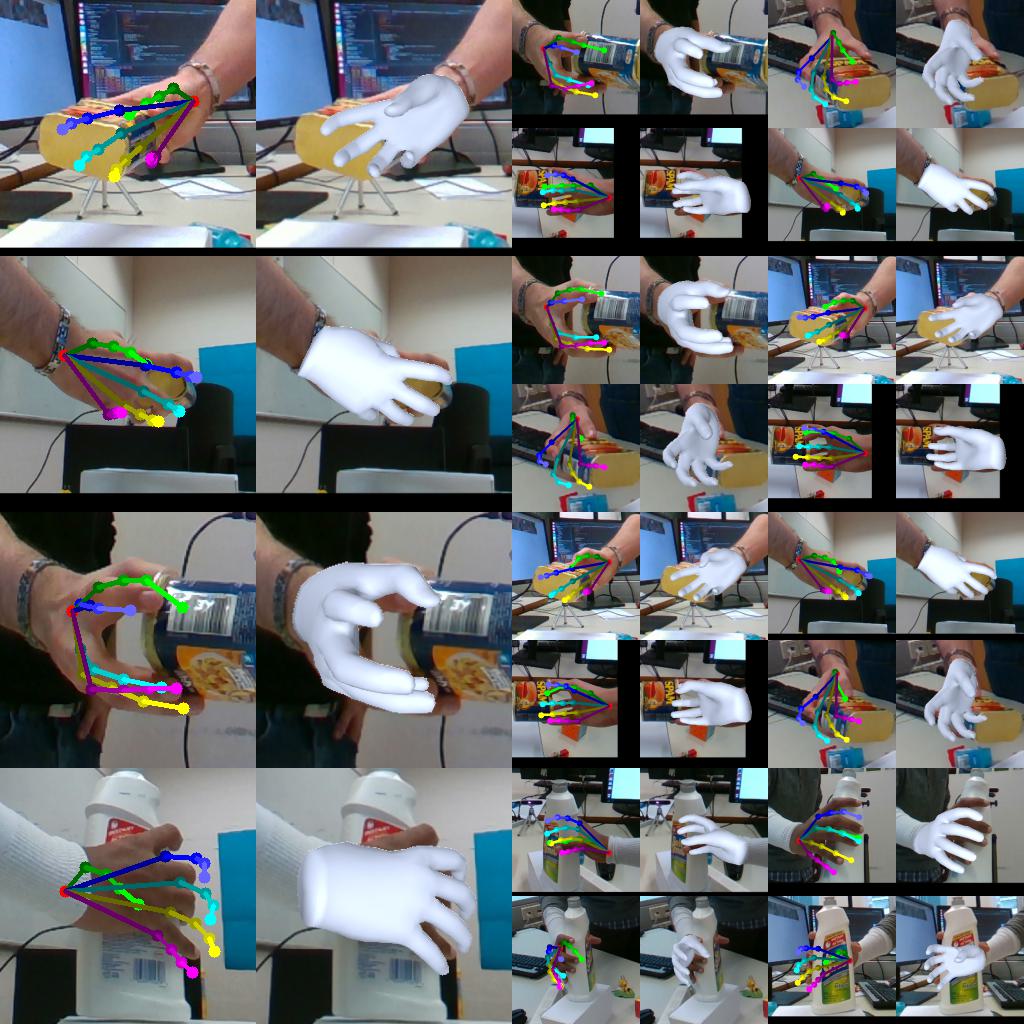}
    \end{subfigure}
    \hfill
    \begin{subfigure}[b]{0.32\textwidth}
        \centering
        \includegraphics[width=\textwidth]{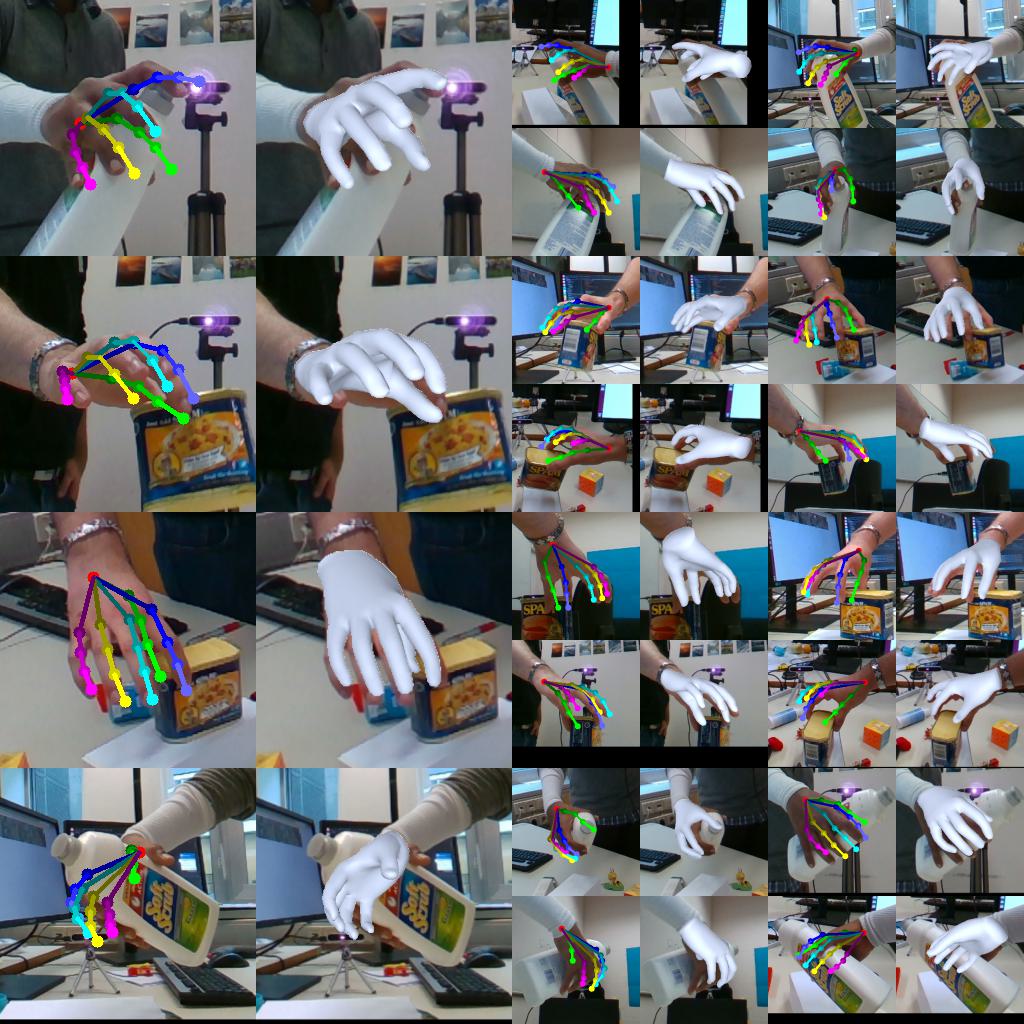}
    \end{subfigure}
    \hfill
    \begin{subfigure}[b]{0.32\textwidth}
        \centering
        \includegraphics[width=\textwidth]{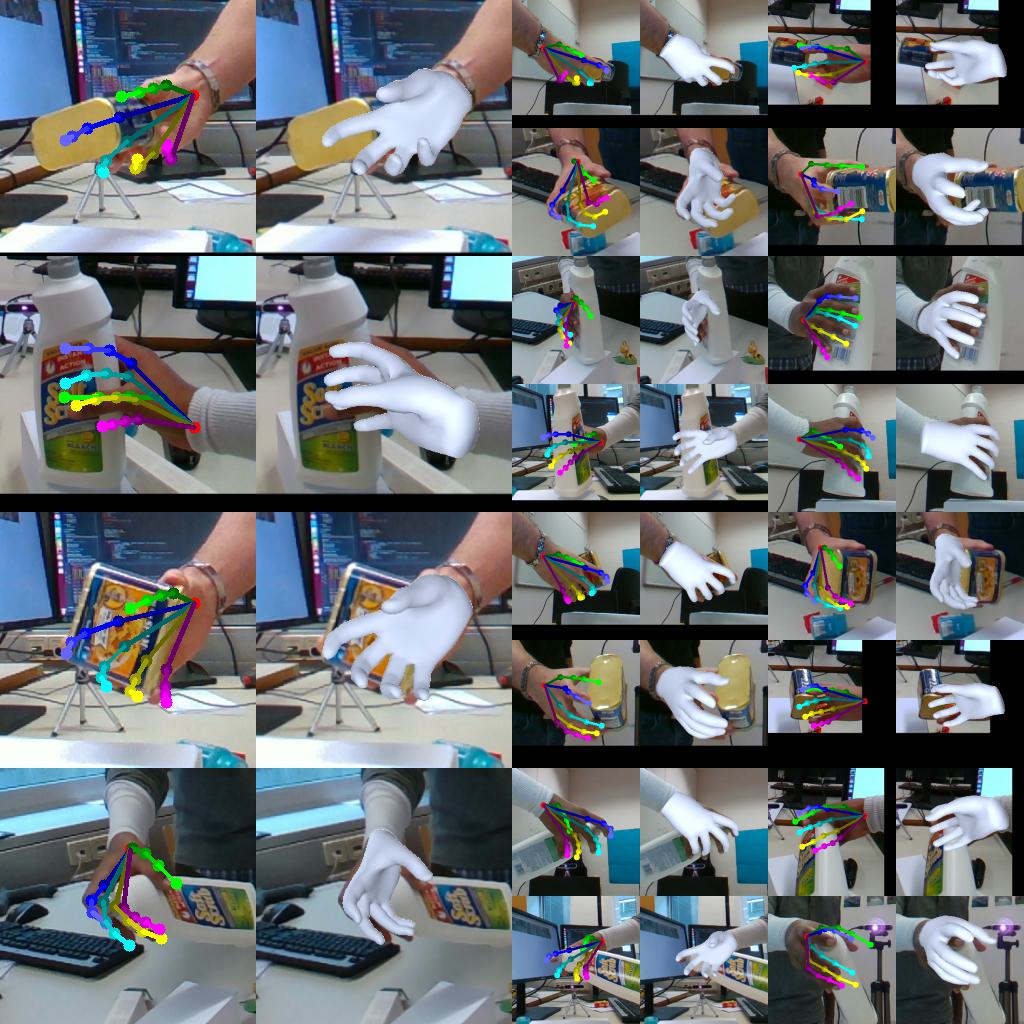}
    \end{subfigure}
    \caption{Qualitative results on the test set of HO3D-MV dataset.}
    \label{fig:ho3d_vis}

\end{figure*}

\begin{figure*}[htb]
    
    \centering
    \begin{subfigure}[b]{0.32\textwidth}
        \centering
        \includegraphics[width=\textwidth]{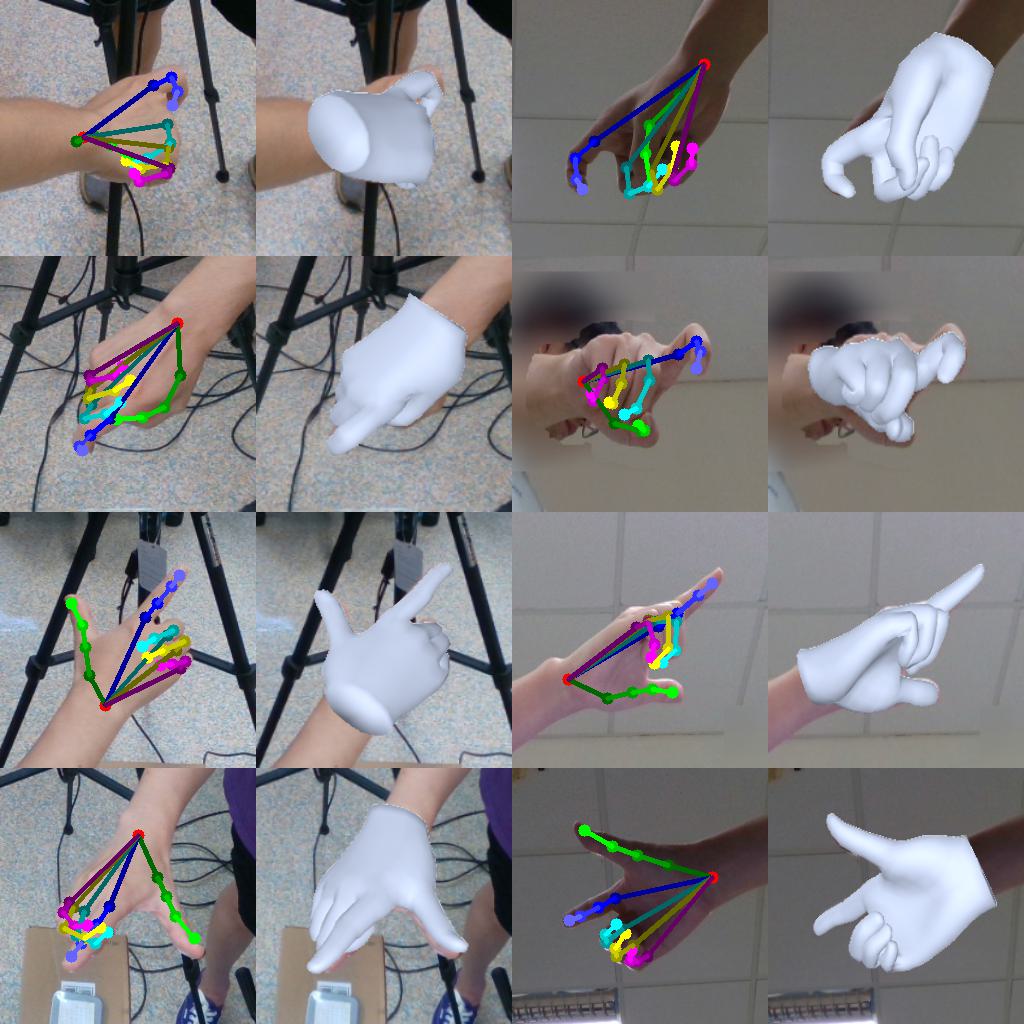}
    \end{subfigure}
    \hfill
    \begin{subfigure}[b]{0.32\textwidth}
        \centering
        \includegraphics[width=\textwidth]{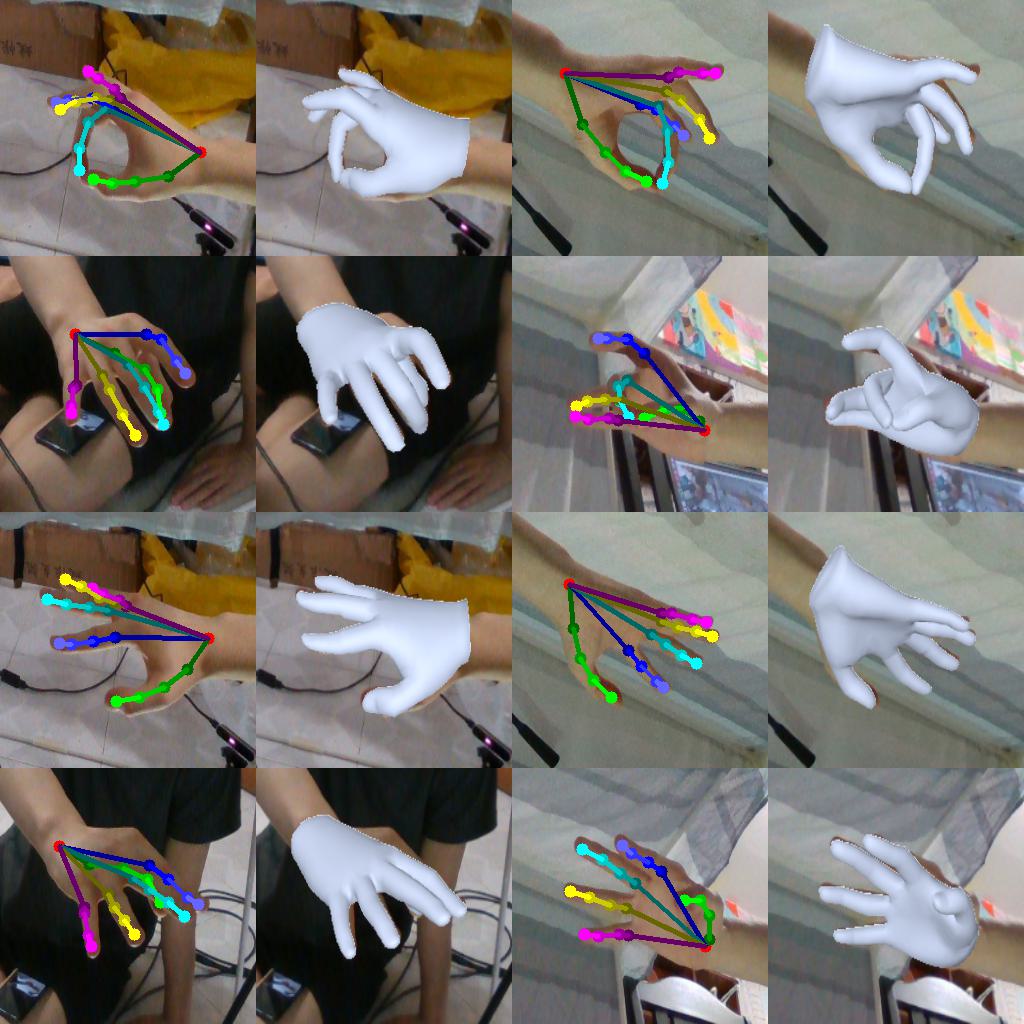}
    \end{subfigure}
    \hfill
    \begin{subfigure}[b]{0.32\textwidth}
        \centering
        \includegraphics[width=\textwidth]{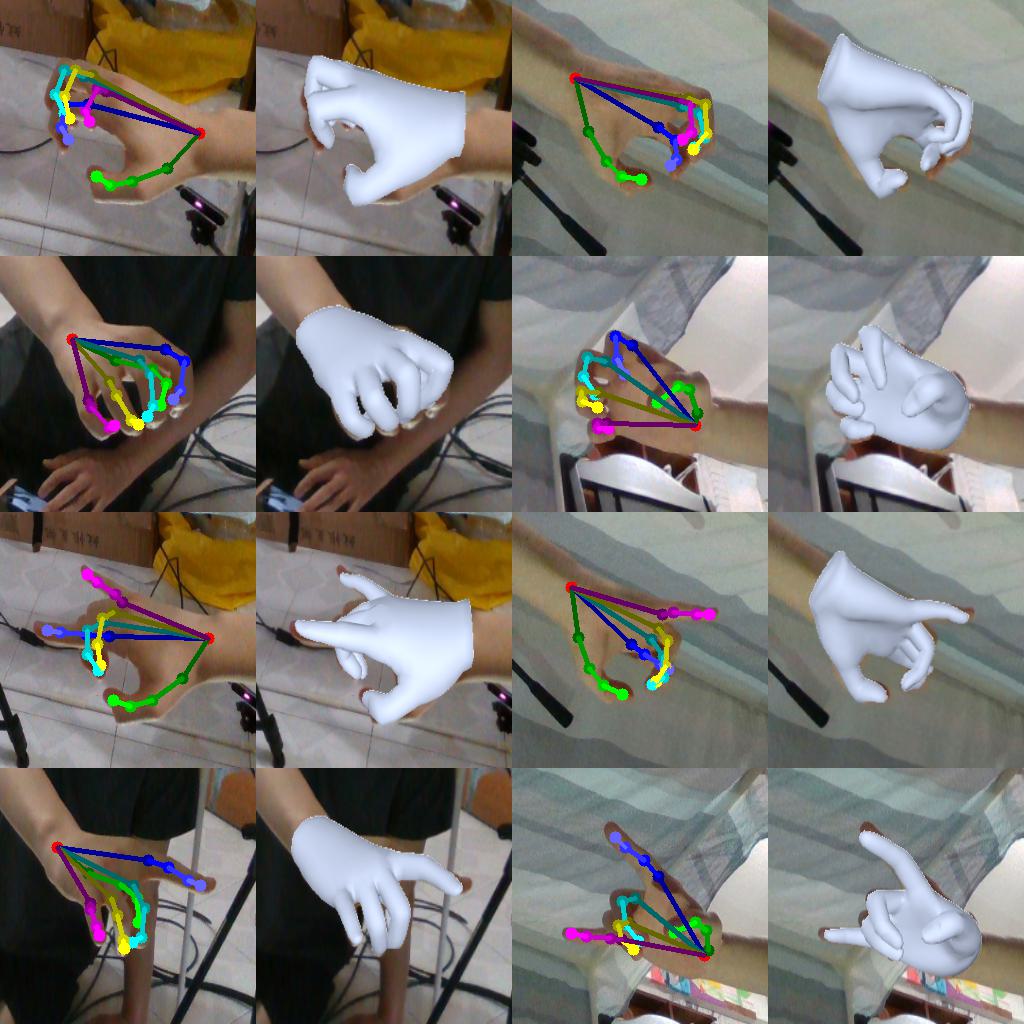}
    \end{subfigure}
    \caption{Qualitative results on the test set of MVHand dataset.}
    \label{fig:mvhand_vis}
\end{figure*}

Table \ref{tab:quantitative} shows that our method consistently outperforms SOTA approaches in terms of MPJPE$_{rel}$ and MPVPE$_{rel}$ across all datasets, while achieving competitive performance in other metrics. In particular, our camera-independent variants, \textbf{``HandMvNet w/o cam."} and \textbf{``HandMvNet-HR w/o cam."}, also show superior performance in most cases. Our method's capacity to implicitly learn 3D geometry demands substantial data, leading to a performance decline on smaller datasets like HO3D-MV as shown in Table \ref{tab:quantitative}. Figure \ref{fig:fps} shows that HandMvNet surpasses other methods in both accuracy (lower MPJPE$_{rel}$) and inference speed (higher FPS). We visualize qualitative results on the DexYCB-MV, HO3D-MV, and MVHand test sets in Figures \ref{fig:dexycb_vis}, \ref{fig:ho3d_vis}, and \ref{fig:mvhand_vis}, respectively.


\subsection{Ablation Study}

\noindent \textbf{Different Backbones.} We compare the results of HandMvNet using ResNet50 as backbone and HandMvNet-HR using HRNet-w40 as backbone in the rows \texttt{6-7, 14-15, 19-20} of Table \ref{tab:quantitative}.

\smallskip \noindent \textbf{Use of Camera Parameters.} In our method, camera parameters are used to add the PE$_{crop}$ positional encoding and loss terms $L_{G2D}$ and $L_{P2D}$. To evaluate the effect of removing camera dependency, we create variants ``HandMvNet w/o cam." and ``HandMvNet-HR w/o cam." by excluding these components. The performance of both versions, with and without camera parameters, are compared in rows \texttt{6-9, 14-17, 19-22} of Table \ref{tab:quantitative}.

\smallskip \noindent \textbf{Impact of Positional Encoding.} In Table \ref{tab:pos-enc}, we examine the effect of different positional encodings on performance. Using the combination of sinusoidal positional encoding ($\text{PE}_{\text{sin}}$), joint-wise encoding ($\text{PE}_{\text{joint}}$)  and crop encoding ($\text{PE}_{\text{crop}}$) results in the best performance.

\smallskip \noindent \textbf{Number of Fusion Layers.} The impact of varying the number of fusion layers is presented in Table \ref{tab:fusion-layers}. We observe that increasing from 3 to 5 layers improves performance, but adding more layers does not further enhance performance, suggesting that 5 layers are optimal.

\begin{figure}[h]
  \centering
  \includegraphics[width=0.44\textwidth]{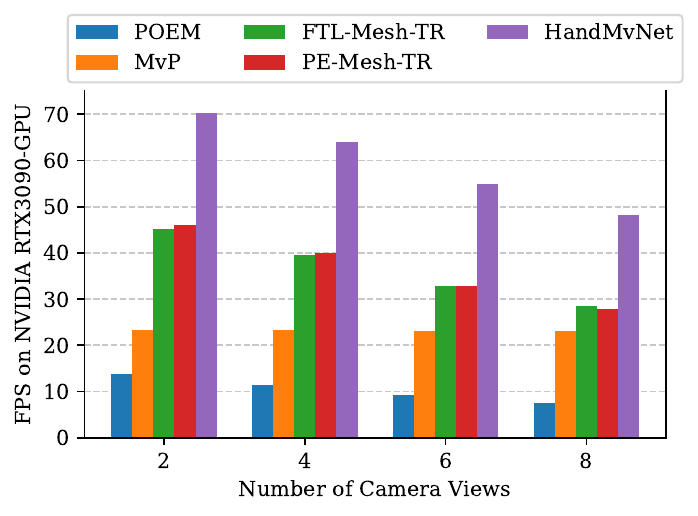}
  \caption{Inference Speed (FPS) comparison across methods with different camera views. HandMvNet achieves the highest FPS across all configurations.}
  \label{fig:fps_vs_cams}
\end{figure}

\smallskip \noindent \textbf{Different Number of Camera Views.} Table \ref{tab:num-cams} shows that model performance improves gradually with increasing the number of camera views. We also compare FPS across different camera views with other approaches in Figure \ref{fig:fps_vs_cams}.

\section{\uppercase{Conclusion}}
\label{sec:discussion}

We introduced HandMvNet, one of the first real-time methods for estimating 3D hand motion and shape from multi-view camera images. Our approach employs a multi-view attention-fusion mechanism that effectively integrates features from multiple viewpoints, delivering consistent and accurate absolute hand poses and shapes, free from the scale-depth ambiguities typically seen in monocular methods. Unlike previous multi-view approaches, HandMvNet eliminates the need for camera parameters to learn 3D geometry. We validated the architecture through extensive ablation studies and compared its performance with state-of-the-art methods. Experiments on public datasets demonstrate the effectiveness of our approach, delivering superior accuracy and inference speed compared to existing methods.

\section*{\uppercase{Acknowledgment}}
\label{sec:ack}
This research has been partially funded by the EU projects FLUENTLY (GA Nr 101058680) and Sharespace (GA Nr 10192889).

\bibliographystyle{apalike}
{\small
    \bibliography{references}
}
\end{document}